\documentclass{article}

\usepackage{arxiv}

\usepackage[utf8]{inputenc} 
\usepackage[T1]{fontenc}    
\usepackage{hyperref}       
\usepackage{url}            
\usepackage{booktabs}      
\usepackage{amsfonts}       
\usepackage{nicefrac}       
\usepackage{microtype}      
\usepackage{lipsum}
\usepackage{graphicx}
\usepackage[ruled,vlined]{algorithm2e}
\usepackage{multirow}
\usepackage{graphicx}
\usepackage{colortbl}
\usepackage{amsmath}
\usepackage{xcolor}
\usepackage{multirow}
\usepackage{makecell}
\usepackage{utfsym}
\usepackage{tabularx}
\usepackage{colortbl}
\graphicspath{ {./images/} }

\title{Denoising-GS: Gaussian Splatting with Spatial-aware Denoising}

\renewcommand{\shorttitle}{DenoisingGS}

\author{
 Qingyuan Zhou \\
  School of Computer Science and Artificial Intelligence\\
  Fudan University\\
  \texttt{zhouqy23@m.fudan.edu.cn}\\
   \And
 Xinyi Liu \\
  School of Computer Science and Artificial Intelligence\\
  Fudan University\\
  \texttt{liuxiny24@m.fudan.edu.cn}\\
  \And
 Weidong Yang\thanks{Corresponding authors} \\
  School of Computer Science and Artificial Intelligence\\
  Fudan University\\
  \texttt{wdyang@fudan.edu.cn}
  \And
 Ning Wang\\
 Software of Engineering\\
Dalian University of Technology\\
    \And
  Shuquan Ye\\
   Department of Information Engineering\\
    The Chinese University of Hong Kong\\
  \And
  Ben Fei\protect\footnotemark[1] \\
  Department of Information Engineering\\
  The Chinese University of Hong Kong\\
  \texttt{benfei@cuhk.edu.hk}
  \And
  Ying He\\
     College of Computing and Data Science\\
 Nanyang Technological University\\
 \texttt{yhe@ntu.edu.sg}
 \And
   Wanli Ouyang\\
  Department of Information Engineering\\
  The Chinese University of Hong Kong\\
 \texttt{wlouyang@cuhk.edu.hk}
}

\date{}

\begin{document}
\maketitle


\begin{abstract}

Recent advances in 3D Gaussian Splatting (3DGS) have achieved significant success in high-fidelity Novel View Synthesis (NVS), yet the optimization process still introduces noisy Gaussian primitives due to the sparse and incomplete initialization from Structure-from-Motion (SfM) point clouds. Most existing methods focus only on adjusting the positions of primitives during optimization, while ignoring the underlying spatial structure. To this end, we introduce a new perspective by formulating the optimization of 3DGS as a primitive denoising process and propose \textbf{Denoising-GS}, a spatial-aware denoising framework for Gaussian primitives by taking both the positions and spatial structure into consideration. Specifically, we design an optimizer that preserves the spatial optimization flow of primitives, facilitating coherent and directed denoising rather than random perturbations. Building upon this, the \textbf{Spatial Gradient-based Denoising} strategy jointly considers the spatial supports of primitives to ensure gradient-consistent updates. Furthermore, the \textbf{Uncertainty-based Denoising} module estimates primitive-wise uncertainty to prune redundant or noisy primitives, while the \textbf{Spatial Coherence Refinement} strategy selectively splits primitives in sparse regions to maintain structural completeness. Experiments conducted on three benchmark datasets demonstrate that Denoising-GS consistently enhances NVS fidelity while maintaining representation compactness, achieving state-of-the-art performance across all benchmarks. Source code and models will be made publicly available.

\end{abstract} 

\section{Introduction}\label{sec:introduction}

Recent progress in 3D scene representation ~\cite{fei20243d,bao20253d,gao2022nerf} has greatly advanced novel view synthesis (NVS), enabling photorealistic rendering from multi-view images. Among these approaches, 3D Gaussian Splatting (3DGS)~\cite{kerbl20233d} has emerged as a leading explicit representation, modeling scenes with anisotropic Gaussian primitives that support real-time rendering and efficient optimization. Owing to its strong balance between rendering fidelity and computational efficiency, 3DGS has inspired numerous extensions across diverse applications, including specular modeling~\cite{song2025reflections,liang2025gus,shi2025gir}, surface reconstruction~\cite{huang20242d,zhou2025mgsr,chen2024pgsr}, SLAM~\cite{matsuki2024gaussian,yan2024gs}, 3D generation~\cite{zhou2024diffgs,chen2024text,chung2023luciddreamer}, and 3D human modeling~\cite{liu2024humangaussian,kocabas2024hugs}.

\begin{figure}[t]
    \centering
    \includegraphics[width=\linewidth]{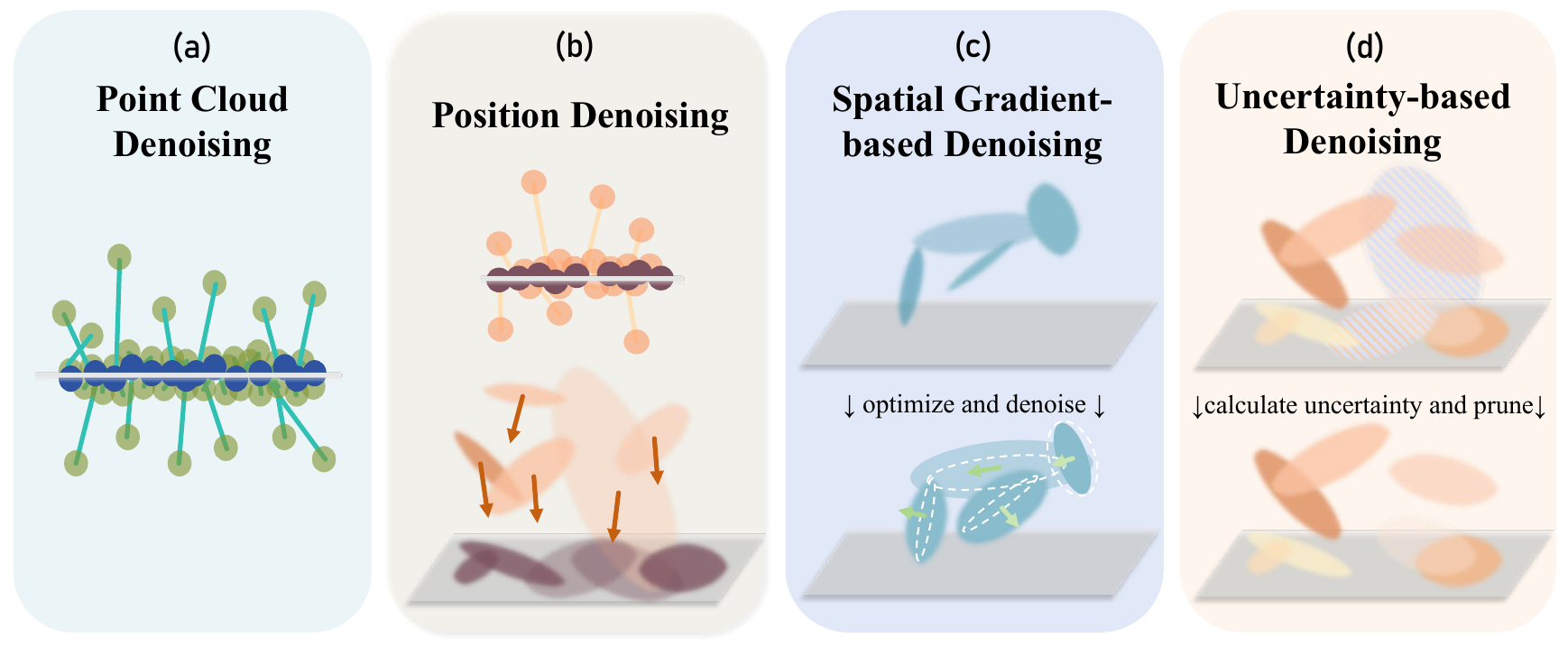}
    \caption{Illustrative comparison of different denoising methods: 
    (a) Point cloud denoising, which moves noisy points toward the GT surface;
    (b) Position denoising of Gaussian primitives, which adjusts the mean positions of noisy primitives toward the GT surface;
    (c) Spatial gradient-based denoising of Gaussian primitives, which optimizes both the mean positions and the spatial structure of primitives based on gradients;
    (d) Uncertainty-based denoising of Gaussian primitives, which estimates the uncertainty of each primitive and prunes those with high uncertainty.}
    \label{fig:teaser}
\end{figure}

Despite its success, 3DGS still faces a fundamental challenge due to its reliance on Structure-from-Motion (SfM) point clouds for initialization. Since these point clouds are often sparse and incomplete, the SGD-based optimization easily becomes trapped in local minima, producing noisy Gaussian primitives that compromise structural fidelity. Recent work, 3DGS-MCMC~\cite{kheradmand20243d}, addresses this issue by introducing opacity-guided relocation and Stochastic Gradient Langevin Dynamics (SGLD)~\cite{welling2011bayesian} perturbations to help primitives escape local minima. 
However, these stochastic updates act only on the mean positions, leaving the covariance (spatial structure) unchanged, and the random perturbations in SGLD often lead to slow convergence, particularly near saddle points or in regions of high curvature~\cite{kim2020s}.

While such stochastic approaches help escape local minima, they do not directly address the noisy nature of the primitives themselves. In practice, the inaccurate initialization and unstable updates in 3DGS generate primitives that behave similarly to noisy points scattered around the true surface. This observation naturally connects 3DGS optimization with the well-studied problem of point cloud denoising. 

In point cloud denoising~\cite{de2023iterativepfn, zhou20253dmambaipf}, noisy points are iteratively pulled toward the ground-truth (GT) surface (\ref{fig:teaser}(a)), while GS-based surface reconstruction methods~\cite{huang20242d,dai2024high,chen2024pgsr,zhou2025mgsr} likewise aim to position primitives near the underlying geometry (\ref{fig:teaser}(b)). 
However, unlike point cloud denoising that involves positional refinement, the positions (means) and spatial structures (covariances) should be simultaneously denoised.
This motivates us to reformulate 3DGS optimization as a Gaussian primitive denoising process, in which both the mean positions and covariance matrices are progressively refined to achieve a coherent, noise-reduced representation. 

Building on this perspective, we propose \textbf{Denoising-GS}, a spatially aware denoising framework that jointly refines the mean and covariance of each primitive. To achieve stable and directed optimization beyond random perturbations, we introduce momentum-biased stochastic exploration into the optimization of 3DGS, which maintains the optimization flow of each primitive, leading to smoother convergence and more consistent updates. 

On top of the momentum-biased stochastic exploration, we design the \textbf{Spatial Gradient-based Denoising} strategy that aligns the spatial updates of primitives with the underlying gradient field (\ref{fig:teaser}(c)), ensuring geometrically consistent refinement. To further address redundancy and incomplete spatial coverage, we incorporate \textbf{Uncertainty-based Denoising}, which leverages uncertainty estimation to identify and prune noisy or redundant primitives (\ref{fig:teaser}(d)). Finally, the \textbf{Spatial Coherence Refinement} strategy replenishes under-represented regions, preserving a compact yet complete scene representation for high-fidelity reconstruction.
We evaluate Denoising-GS on three public datasets (Mip-NeRF 360~\cite{barron2022mip}, Tanks and Temples~\cite{knapitsch2017tanks}, and Deep Blending~\cite{hedman2018deep}), and demonstrate state-of-the-art results among all baselines. Our contributions are fourfold:
\begin{itemize}

\item We reformulate the GS optimizing procedure as a Gaussian primitive denoising process for achieving accurate scene representation, and propose a spatial-aware 3DGS denoising framework.
\item We introduce the momentum-biased stochastic exploration to enable primitives to stochastically explore global optimal positions while maintaining the optimization flow of the spatial process.
\item We design the \textbf{Spatial Gradient-based Denoising} module that considers the spatial support of each primitive, ensuring its spatial updates remain consistent with the underlying gradient field.
\item We leverage uncertainty estimation and design the \textbf{Uncertainty-based Denoising} module to identify and prune redundant or noisy primitives, and introduce the \textbf{Spatial Coherence Refinement} strategy for compensation of sparse regions.
\end{itemize}

\section{Related Work}\label{sec:related}

\textbf{GS-based Novel View Synthesis.}
The advent of 3DGS has fundamentally redefined the landscape of NVS and has inspired numerous subsequent studies~\cite{fei20243d, wu2024recent} that aim to further enhance rendering fidelity, convergence speed, and computational efficiency.
These methods address issues such as removing floating artifacts and achieving alias-free rendering~\cite{yu2024mip,yan2024multi,steiner2025aaa}, refining the densification and pruning strategies of Gaussian primitives~\cite{cheng2024gaussianpro,kheradmand20243d,yang2025improving,zhang2024pixel,lyu2025resgs,ye2024absgs}.
Meanwhile, other approaches aim to balance rendering quality and efficiency through budget-aware optimization~\cite{mallick2024taming,hanson2025speedy}, anchor-based Gaussian Splatting~\cite{lu2024scaffold,zhang2025sogs,xie2025hash}, improved Gaussian rasterization~\cite{feng2025flashgs,radl2024stopthepop,bulo2025hardware}, and alternative primitive (kernel) representations ~\cite{hamdi2024ges,li20253d,zhu20253d}.

\textbf{Optimizing Gaussian Primitives in Spatial.}
Most existing GS-based methods follow an SGD-based optimizer to iteratively refine Gaussian primitives.
In terms of geometric fidelity, GS-based surface reconstruction methods~\cite{huang20242d,chen2024pgsr,zhou2025mgsr,dai2024high} flatten the primitives in spatial to produce scene representations that better align with GT mesh surfaces.
To further accelerate optimization, 3DGS-LM~\cite{hollein20253dgs} introduces a Levenberg-Marquardt optimizer, achieving comparable rendering quality with faster speed.
Meanwhile, 3DGS-MCMC~\cite{kheradmand20243d} adopts SGLD to introduce stochastic explorations on the mean positions during optimization, which helps prevent local optima and improve rendering robustness.
In contrast to these approaches, our Denoising-GS preserves the spatial optimization trajectory of primitives in stochastic exploration, and models the spatial relationship between the mean and scaling, leveraging gradients to denoise primitives. 
It is also worth noting that Pixel-GS~\cite{zhang2024pixel} and AbsGS~\cite{ye2024absgs} address a similar issue to our Spatial Gradient-based Denoising module, namely the influence of gradients.
However, their main claims focus on utilizing gradients to facilitate the growth of large Gaussian primitives or to split oversized ones, without explicitly considering the spatial interaction and coordinated updates between the gradients of the mean and the scaling.


\textbf{Uncertainty Estimation in GS.}
Recent works have explored uncertainty estimation within Gaussian Splatting representations, covering joint optimization of per-primitive uncertainty~\cite{li2024variational}, next-best-view selection~\cite{jiang2024fisherrf}, and various downstream applications~\cite{huang2024gaussianmarker,hanson2025pup}.
FisherRF~\cite{jiang2024fisherrf} and PUP 3D-GS~\cite{hanson2025pup} both use Fisher Information to estimate uncertainties in GS representations, similar to our Denoising-GS.
However, FisherRF focuses on actively selecting informative training views based on uncertainty estimation, while PUP 3D-GS performs post-hoc uncertainty estimation to prune redundant Gaussians for downstream tasks that require sparse scene representations.
In contrast, Denoising-GS integrates uncertainty estimation directly into the primitive denoising process, thereby achieving a tighter and more efficient scene representation.

\section{Preliminaries}
\label{sec:preliminaries}

\textbf{3DGS}~\cite{kerbl20233d} is an explicit scene representation that utilizes a set of 3D anisotropic primitives to model a radiance field.
Each of the primitives is represented by a 3D Gaussian distribution, initialized from SfM point clouds, as illustrated in \ref{eq:p1},
\begin{equation}
    G=e^{-\frac{1}{2} \mathcal{M}^\top \Sigma^{-1} \mathcal{M}},
    \label{eq:p1}
\end{equation}
where $\mathcal{M} \in \mathbb{R}^3$ denotes the mean of the Gaussian distribution, $\Sigma$ stands for the covariance matrix.
Considering the inherent physical significance of the covariance matrix, which challenges to update conventionally in back-propagation, the covariance matrix is decomposed into rotation matrix $\mathbf{R}$ and scaling matrix $\mathbf{S}$ by \ref{eq:p2},
\begin{equation}
    \Sigma=\mathbf{R S S}^\top \mathbf{R}^\top,
    \label{eq:p2}
\end{equation}
where $\mathbf{R}$ represents quaternions, and $\mathbf{S}$ denotes to scaling factors.

\textbf{3DGS-MCMC}~\cite{kheradmand20243d} reinterprets the training process of 3DGS as a Markov Chain Monte Carlo (MCMC) sampling procedure. Instead of relying on heuristic densification or pruning strategies, it views the set of Gaussian primitives as random samples drawn from an underlying probability distribution that characterizes the physical scene. Under this probabilistic perspective, the conventional gradient-based optimization of 3DGS is reformulated as a Stochastic Gradient Langevin Dynamics (SGLD) update, shown in \ref{eq:p5},
\begin{equation}
\underbrace{\mathcal{M} \leftarrow \mathcal{M} - \lambda_{\mathrm{lr}} \nabla\mathcal{M} L_{\mathrm{total}}}_{\text{vanilla 3DGS update}}+\lambda_{\mathrm{noise}}\lambda_{\mathrm{lr}} \boldsymbol{\xi},
\label{eq:p5}
\end{equation}
which is derived by augmenting the Adam-based parameter update of vanilla 3DGS with an additional SGLD noise term to enable stochastic exploration in the parameter space. Here, $L_{\mathrm{total}}$ denotes total loss function, $\lambda_{\mathrm{noise}}$ controls the magnitude of the injected noise, and $\boldsymbol{\xi} \sim \mathcal{N}(0, \mathbf{I})$ denotes Gaussian noise.

Additionally, 3DGS-MCMC introduces a novel strategy for densifying and pruning Gaussian primitives. In practice, primitives with low opacity are relocated to the positions of primitives with high opacity, following the transformations defined in \ref{eq:p6} and \ref{eq:p7}.
\begin{equation}
o^{\mathrm{new}}_{1,\dots,N} = 1 - \sqrt[N]{\,1 - o^{\mathrm{old}}_{N}\,},
\label{eq:p6}
\end{equation}
\begin{equation}
\Sigma^{\mathrm{new}}_{1,\dots,N} = \left(o^{\mathrm{old}}_{N}\right)^{2}
\left(
\sum_{i=1}^{N} \sum_{k=0}^{i-1}
\binom{i-1}{k}
\frac{(-1)^{k}\left(o^{\mathrm{new}}_{N}\right)^{k+1}}{\sqrt{k+1}}
\right)^{-2}
\Sigma^{\mathrm{old}}_{N},
\label{eq:p7}
\end{equation}
where $o$ is the opacity, $\Sigma$ denotes the covariance matrix, and $N$ indicates the number of new primitives cloning at the same position.

\section{Method}\label{sec:Method}

\subsection{Overview}

Denoising-GS redefines the optimization of 3DGS as a denoising process of Gaussian primitives.
Specifically, we first introduce the momentum-biased stochastic exploration into 3DGS, which preserves the spatial optimization trajectory of primitives and enables more stable and direction-aware denoising (\ref{sec:4.2}). 
To address the inefficiency issue where the translation of the mean is counteracted by concurrent scaling along the same axis, we design a spatial denoising strategy that jointly considers the gradients of means and scalings to align spatial variations with gradient directions (\ref{sec:4.3}). 
Furthermore, we introduce an uncertainty estimation module to identify and prune redundant or noisy primitives (\ref{sec:4.4}), and the spatial coherence refinement strategy to compensate for sparse regions (\ref{sec:4.5}). 

\begin{figure*}[t]
    \centering
    \includegraphics[width=\linewidth]{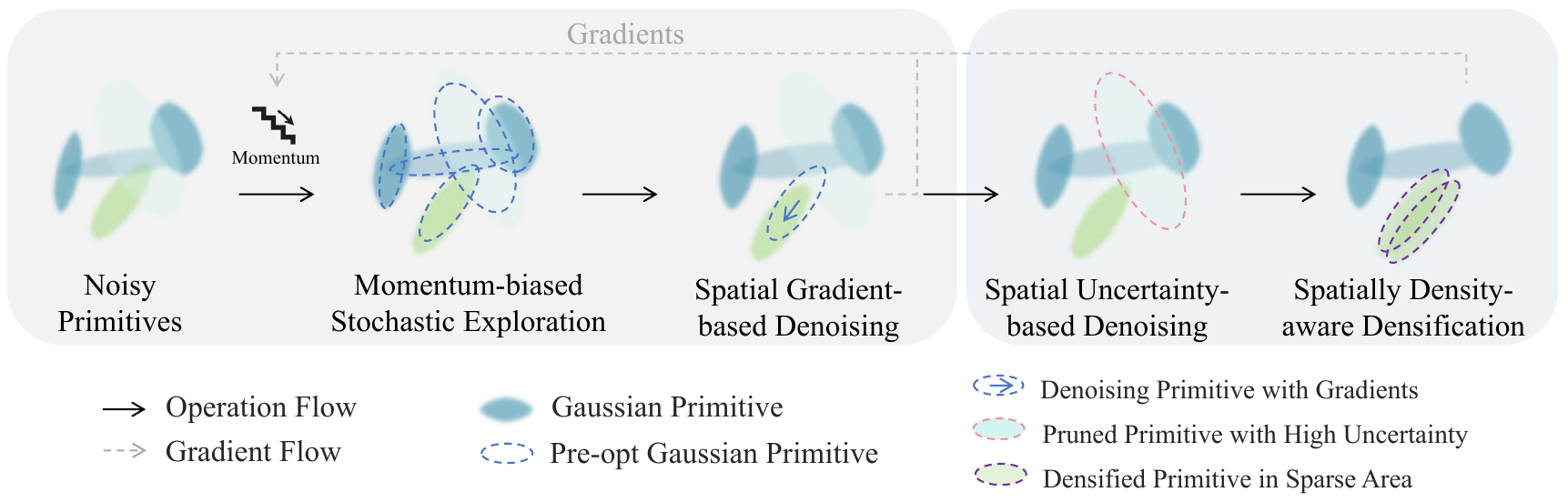}
    \caption{Overview of the proposed Denoising-GS framework. The pipeline starts from noisy SfM-initialized Gaussian primitives and progressively refines them through a multi-stage denoising process, yielding a compact and accurate representation for high-fidelity NVS. The proposed momentum-biased exploration preserves the spatial optimization trajectory of primitives, ensuring that the denoising exploration is coherent and directed rather than purely stochastic. The spatial gradient-based denoising module jointly considers the spatial support of primitives, and performs denoising guided by the gradients. Furthermore, the uncertainty-based denoising module prunes redundant or noisy primitives, while the spatial coherence refinement strategy compensates for sparse regions to maintain completeness and fidelity.}
    \label{fig:method}
\end{figure*}

\subsection{Momentum-biased Stochastic Exploration}
\label{sec:4.2}
3DGS-MCMC~\cite{kheradmand20243d} introduces SGLD~\cite{welling2011bayesian} perturbations to inject random noise into each Gaussian primitive during optimization, allowing exploration of more potential positions and alleviating the issue of local minima encountered in standard SGD. 
However, SGLD often results in unstable trajectories and slow convergence.
To address these issues, we adopt the Momentum-biased Stochastic Exploration~\cite{kim2020s} strategy to 3DGS. 
In contrast to the vanilla SGLD, which updates parameters solely based on instantaneous gradients and stochastic noise, we introduce an exponential moving average of historical gradients (momentum) into the stochastic exploration, effectively stabilizing the stochastic updates while preserving the asymptotic correctness of the sampling process.

Formally, given the mean of a Gaussian primitive $\mathcal{M}_t \in \mathbb{R}^3$ at iteration $t$ in optimization, the SGLD update is realized by introducing a drift component. The update rule is defined as:
\begin{equation}
\mathcal{M}_{t+1} = \mathcal{M}_t 
 - \lambda_{t} \nabla \mathcal{M}_t L_{\mathrm{total}}
 + \sqrt{2 \lambda_{t} \tau} \, \boldsymbol{\xi}_t,
 \label{eq:m1}
 \end{equation}
 where $\nabla \mathcal{M}_t$ is the gradient of $\mathcal{M}_t$,
 $\lambda_{t}$ is the learning rate,
 $\tau$ is the temperature coefficient controlling the stochasticity,
and $\boldsymbol{\xi}_t \sim \mathcal{N}(0, \mathbf{I})$ is Gaussian noise.
3DGS-MCMC simplifies this formula to \ref{eq:p5}.

To alleviate the aforementioned issues and endow the stochastic noise with a memory bank, our strategy incorporates a momentum term $\mathbf{m}_t$ which maintains an exponentially decayed history of past gradients to guide the exploration.
\begin{equation}
 \mathbf{m}_{t} = \beta_1 \mathbf{m}_{t-1} + (1 - \beta_1)\nabla \mathcal{M}_{t-1},
 \label{eq:m2}
\end{equation}
where $\nabla \mathcal{M}_{t-1}$
is the gradient of the mean $\mathcal{M}$ of the Gaussian primitive, and $\beta_1$ is a smoothness controlling coefficient. 
Our parameter update is then formulated as:
 \begin{equation}
\mathcal{M}_{t+1} = \mathcal{M}_t 
 - \lambda_{t} \nabla \mathcal{M}_t L_{\mathrm{total}}
 + \sqrt{2 \lambda_{t} \tau} \, \boldsymbol{\xi}_t -\alpha\lambda_{t}\mathbf{m}_{t},
 \label{eq:msgld}
\end{equation}
 where $\alpha\in(0,1]$
 is a bias factor that controls the relative contribution of the momentum term in the drift component.

In the implementation, the magnitude of the Gaussian noise disturbed to the means is opacity-aware and adapted to the spatial shape of each Gaussian primitive, whereas the momentum term serves purely as an offset determined by historical gradients, independent of both opacity and the covariance matrix. The detailed implementation is provided in the Appendix.

\subsection{Spatial Gradient-based Denoising}
\label{sec:4.3}

Although SGLD and our momentum-biased stochastic exploration could eliminate the positional noise of Gaussian primitives, the spatial noise should also be taken into consideration.
In other words, the mean position $\mathcal{M}$ and the covariance matrix $\boldsymbol{\Sigma}$ of each Gaussian primitive should be jointly denoised, as the latter defines the effective spatial support of the primitive.

\begin{figure}[t]
    \centering
    \small
    \includegraphics[width=0.7\linewidth]{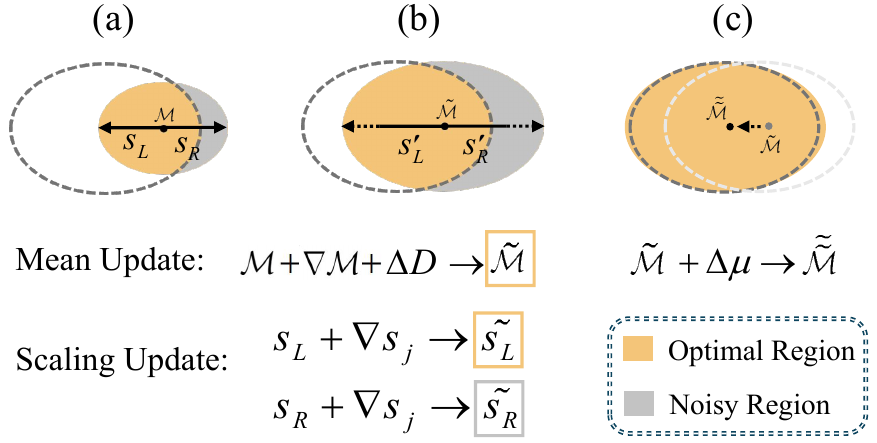}
    \caption{A simplified illustration of the Spatial Gradient-based Denoising process. 
    Note that the diagram represents an idealized case, and we assume that all Gaussian primitives move only along the horizontal direction, with rotation neglected.
    $\mathcal{M}$ is the mean position of Gaussian primitive, $\nabla\mathcal{M}$ is gradient of $\mathcal{M}$, $\Delta D$ is the momentum-biased stochastic exploration component, $s_L = s_R$ represent the half-lengths of the primitive’s major axis $j$, and $\nabla s_j$ is the scaling gradient on $j$ axis. $\tilde{\mathcal{M}}$ is the mean position after the momentum-biased stochastic exploration. $\tilde{\tilde{\mathcal{M}}}$ is the mean position of the denoised primitive, and $\Delta \mu$ is the denoising term.
    }
    \label{fig:denoise}
\end{figure}

During the iterative optimization of a Gaussian primitive, the gradients of the mean and the covariance (decomposed into scaling and rotation) are computed independently based on the total loss.
However, in specific cases, the movement of the mean may align with the primitive’s expansion or contraction, causing redundant updates and thus reducing the effectiveness of optimization.
A simplified illustration is shown in \ref{fig:denoise}, which represents an idealized case where rotation $R = I$. 
The dashed Gaussian primitive denotes the optimal region, and our objective is to iteratively optimize the yellow primitive to cover this target region.
Primitives that lie outside the target region (colored in gray) are considered part of the noisy region.
\ref{fig:denoise}(a) shows the initial state, while \ref{fig:denoise}(b) presents the state after the optimization of the momentum-biased stochastic exploration. 
During this process, the mean $\mathcal{M}$ shifts leftward under the guidance of gradient $\nabla \mathcal{M}$ and the exploration induced in \ref{sec:4.2} ($\Delta D$). Meanwhile, both $s_L$ and $s_R$ expand synchronously with the gradient $\nabla s_j$.
This gradient-guided update, however, also enlarges the gray region, which corresponds to the noisy region.
To address this issue, the proposed Spatial Gradient-based Denoising module introduces a denoising term $\Delta \mu$ (\ref{fig:denoise}(c)) that adaptively adjusts both the means and covariance matrices, ensuring that the Gaussian primitive is denoised toward its optimal position.

However, in a practical scenario, the Gaussian primitives no longer move strictly along the coordinate axes and may undergo rotations, resulting in directly computing the gradient of the covariance matrix $\boldsymbol{\Sigma}$ is computationally expensive.
Following the standard parameterization in~\cite{kerbl20233d}, we define the means as $\mathcal{M} = (\mathcal{M}_{x}, \mathcal{M}_{y}, \mathcal{M}_{z})$, and represent the covariance matrix $\boldsymbol{\Sigma}$ as defined in \ref{eq:p2}, where the scaling is $\mathbf{S} = \mathrm{diag}(s_x, s_y, s_z)$ and the rotation is $\mathbf{R}$.
At each iteration during optimization, the gradients of mean $\nabla \mathcal{M} = (\nabla \mathcal{M}_x, \nabla \mathcal{M}_y, \nabla \mathcal{M}_z)$ indicate the desired directional movement of the means. The scaling gradients $\nabla \mathbf{S} = (\nabla \mathbf{S}_x, \nabla \mathbf{S}_y, \nabla \mathbf{S}_z)$ correspond to the expansion or contraction along the principal axes. 
To ensure consistent updates, we consider both the mean and its gradients within the same local coordinate system of each Gaussian.
Specifically, we first transform the mean gradients into the local coordinate system of the Gaussian via:
\begin{equation}
\nabla \mathcal{M}^{G} = \mathbf{R}^\top \nabla \mathcal{M},
\label{eq:m3}
\end{equation}
where $\nabla \mathcal{M}^{G}$ denotes the mean gradients in local Gaussian coordinate system, $\mathbf{R}$ is rotation matrix. 
Note that although the rotation is stored as a quaternion in the GS pipeline, the corresponding $\mathbf{R}$ is expressed as a $3 \times 3$ matrix.
We then identify the principal axes of both the mean gradients $\nabla \mathcal{M}^{G}$ and scaling gradients $\nabla \mathbf{S}$ by locating their maximum components.
If the two axes are aligned, a directional denoise term $\Delta \mu$ is added along the same axis on $\mathcal{M}$ to prevent redundant movement:
\begin{equation}
\Delta \mu=|\nabla \mathbf{S}_j| \cdot \mathrm{sign}(\nabla \mathcal{M}_{j}^{G}).
\label{eq:m4}
\end{equation}
Here, $\nabla \mathbf{S}_j$ and $\nabla \mathcal{M}^{G}_j$ denote the scaling and mean gradient values along the $j$-th principal axis in the local Gaussian coordinate system, where $j$ corresponds to one of the $x$, $y$, or $z$ axes. The \textit{sign} function preserves the direction of the mean gradient, ensuring that the denoising displacement occurs in the same directional sense as the original gradient update.
Finally, the means of primitives are updated as:
 \begin{equation}
\mathcal{M}_{t+1} = \mathcal{M}_t 
 - \lambda_{t} \nabla \mathcal{M}_t L_{\mathrm{total}}
 +\Delta D +\beta_2\mathbf{R}\Delta \mu,
 \label{eq:m5}
\end{equation}
where $\Delta D = \sqrt{2 \lambda_{t} \tau} \, \boldsymbol{\xi}_t -a\lambda_{t}\mathbf{m}_{t} $ comes from the momentum-biased stochastic exploration defined in \ref{eq:msgld}, with consistent variable names.
$\beta_2\in(0,1]$ is the controlling coefficient of the denoising term $\Delta \mu$.

\subsection{Uncertainty-based Denoising}
\label{sec:4.4}
To achieve a more efficient and compact scene representation, we present a strategy that leverages uncertainty estimation to prune primitives and enable denoising.
Inspired by~\cite{jiang2024fisherrf,hanson2025pup}, we utilize Fisher Information to compute the primitive uncertainty based on means and scalings. Specifically, Fisher Information measures the sensitivity of model parameters to the observed data and can be defined as the variance of the score of the log-likelihood:
\begin{equation}
\mathbf{I}(\theta) = \mathbb{E}_{X \sim p(X \mid \theta)} 
\left[ \nabla_{\theta} \log p(X \mid \theta) \, 
\nabla_{\theta} \log p(X \mid \theta)^\top \right],
\label{eq:f1}
\end{equation}
where  $\theta$ denotes the model parameters,  $X$ is the observed data, and $p(X \mid \theta)$ is the likelihood. 

However, direct computation of this expectation is intractable in the Gaussian Splatting pipeline. 
In our Denoising-GS framework, we approximate the Fisher information for each Gaussian primitive using the gradients of the L2 rendering loss with respect to the primitive’s parameters $\theta$, including the $\mathcal{M}$ and scalings $\mathbf{S}$.
For a single primitive, the Fisher information matrix is approximated via the outer product of the gradient:
\begin{equation}
F_i = \sum_{v \in \mathcal{V}} \nabla{\theta_i} \mathcal{L}_v \, (\nabla{\theta_i} \mathcal{L}_v)^\top,
\label{eq:f2}
\end{equation}
where $\nabla{\theta_i}$ is the gradient of parameters $\theta_i$, and $\mathcal{V}$ denotes the set of all camera views, 
and $\mathcal{L}_v = \| I_v - \hat{I}_v(\theta) \|_2^2$ is the L2 rendering loss of view $v$, where $I_v$ is the GT image and $\hat{I}_v(\theta)$ is the rendered image. 
The uncertainty of the $i$-th primitive is calculated by:
\begin{equation}
u_i = \sum_{j=1}^{6} \sigma_j(F_i),
\label{eq:f3}
\end{equation}
where $ u_i $ denotes the uncertainty of the $i$-th Gaussian primitive, $ F_i $ represents the approximate Fisher information matrix computed for the parameters of the $ i $-th primitive, and $ \sigma_j(F_i) $ is the $ j $-th singular value of $ F_i $ obtained through singular value decomposition (SVD).
The summation runs over six singular values, corresponding to the three dimensional $\mathcal{M}$ and three dimensional $\mathbf{S}$.
The summation over all singular values corresponds to the total information magnitude of the primitive in the spatial domain. We interpret the uncertainty $u$ as an inverse measure of the information contained in each primitive.
Therefore, primitives with smaller $u$ values are considered to have higher uncertainty and lower confidence,
which typically correspond to noisy or redundant points. These primitives are identified and pruned during this denoising stage.
The implementation is provided in the Appendix.

\subsection{Spatial Coherence Refinement}
\label{sec:4.5}
To achieve a more expressive scene representation and improve rendering fidelity, we introduce a spatial coherence refinement strategy that exploits spatial density to adaptively guide the selective densification of Gaussian primitives in sparse regions, thereby enhancing scene details while maintaining representation compactness.
Specifically, we compute the distance between the center of each primitive and its \(k\) nearest neighbors, defined as:
\begin{equation}
d_i = \frac{1}{k} \sum_{j=1}^{k} \| x_i - x_{i,j} \|_2^2,
\label{eq:4.1}
\end{equation}
where \(x_i\) denotes the position of the \(i\)-th primitive, and \(x_{i,j}\) represents the position of one of the \(k\)-th nearest neighbors. Primitives with larger $d$ are considered to lie in sparse regions and are densified according to \ref{eq:p6} and \ref{eq:p7}. The detailed implementation is provided in the Appendix.

\section{Experiments}\label{sec:Experiments}

\begin{table*}[t]
\caption{Results of two baselines and eight GS-based NVS methods on the Mip-NeRF 360, Tank and Temples, and Deep Blending datasets. 
The \textcolor{red!60}{red}, \textcolor{orange!60}{orange} and \textcolor{yellow}{yellow} colors represent the top three results. 
}
\centering
\resizebox{\linewidth}{!}{
\begin{tabular}{c|r|r|ccc|ccc|ccc}
\toprule [1pt]
\multirow{2}{*}{\#}&\multirow{2}{*}{Method}&\multirow{2}{*}{Venue}&\multicolumn{3}{c|}{Mip-NeRF 360}&\multicolumn{3}{c|}{Tanks and Temples}&\multicolumn{3}{c}{Deep Blending}\\

&&&SSIM$\uparrow$&PSNR$\uparrow$&LPIPS$\downarrow$&SSIM$\uparrow$&PSNR$\uparrow$&LPIPS$\downarrow$&SSIM$\uparrow$&PSNR$\uparrow$&LPIPS$\downarrow$\\
\midrule
\multirow{2}{*}{(a)}&Vanilla 3DGS~\cite{kerbl20233d}&ToG'23&0.880&29.258&0.168&0.845&23.649&0.178&\cellcolor{yellow!40}0.900&29.475&0.247\\

&3DGS-MCMC~\cite{kheradmand20243d}&NeurIPS'24&\cellcolor{orange!40}0.896&\cellcolor{orange!40}29.893&0.186&\cellcolor{orange!40}0.860&\cellcolor{yellow!40}24.290&0.190&0.895&\cellcolor{yellow!40}29.665&0.320\\
\midrule

\multirow{8}{*}{(b)}&\scalebox{0.8}[1]{Variational-3DGS}~\cite{li2024variational}&NeurIPS'24&0.871&28.744&0.185&0.839&23.324&0.184&0.890&28.732&0.265\\

&Taming 3DGS~\cite{mallick2024taming}& \scalebox{0.7}[1]{SIGGRAPH Asia'24}&0.862&28.930&0.204&0.833&23.854&0.213&\cellcolor{orange!40}0.901&\cellcolor{red!40}29.968&0.273\\


&Mip-Splatting~\cite{yu2024mip}&CVPR'24&\cellcolor{yellow!40}0.889&29.456&\cellcolor{orange!40}0.149&0.856&23.768&\cellcolor{yellow!40}0.158&0.898&29.310&\cellcolor{orange!40}0.242\\

&GES~\cite{hamdi2024ges}&CVPR'24&0.865&28.766&0.196&0.837&23.470&0.197&\cellcolor{orange!40}0.901&29.553&0.252\\

&AbsGS~\cite{ye2024absgs}&ACM MM'24&0.859&28.645&0.185&0.852&23.653&0.163&\cellcolor{orange!40}0.901&29.612&\cellcolor{red!40}0.239\\

&Pixel-GS~\cite{zhang2024pixel}&ECCV'24&0.886&29.454&\cellcolor{yellow!40}0.159&0.853&23.730&\cellcolor{orange!40}0.151&0.892&28.858&0.251\\

&GS-LPM~\cite{yang2025improving}&CVPR'25&0.887&29.378&0.160&0.845&23.880&0.186&\cellcolor{yellow!40}0.900&29.421&\cellcolor{yellow!40}0.246\\

&3D-HGS~\cite{li20253d}&CVPR'25&0.881&\cellcolor{yellow!40}29.772&0.162&\cellcolor{yellow!40}0.857&\cellcolor{orange!40}24.395&0.165&\cellcolor{orange!40}0.901&29.554&0.248\\
\midrule

(c)&\textbf{Denoising-GS} &\textbf{Ours}&\cellcolor{red!40}0.897&\cellcolor{red!40}29.925&\cellcolor{red!40}0.142&\cellcolor{red!40}0.868&\cellcolor{red!40}24.561&\cellcolor{red!40}0.149&\cellcolor{red!40}0.906&\cellcolor{orange!40}29.872&\cellcolor{red!40}0.239\\
\toprule [1pt]
\end{tabular}
}
\label{tb:exp_1}
\end{table*}

\subsection{Implementation}
\label{sec:5.1}
Our Denoising-GS framework is built using PyTorch 2.5.1 and CUDA 12.4. All experiments are conducted on an NVIDIA H100 GPU. 
For all experimental scenes, $\alpha$ is 0.05, $\beta_1$ is 0.9, $\beta_2$ is 0.5, and $k$ is 3. In the Uncertainty-based Denoising stage, we remove the top 10\% of the most uncertain primitives, while conducting Spatial Coherence Refinement a total of five times, where each iteration divides the 0.05\% of the most spatially sparse primitives. 
These components will be ablated in \ref{sec:ablations}.

\textbf{Datasets and Metrics.} 
Following the experimental setup of prior work~\cite{kheradmand20243d}, Denoising-GS is evaluated on eleven scenes from three datasets, including seven scenes from Mip-NeRF 360~\cite{barron2022mip}, two outdoor scenes from Tanks and Temples~\cite{knapitsch2017tanks}, and two indoor scenes from Deep Blending~\cite{hedman2018deep}. 
We explore the use of denser and noisier input SfM point clouds on Mip-NeRF 360, which is optimized for 35K iterations.
Tanks and Temples, and Deep Blending are optimized for 30K iterations.
Structural Similarity Index Metric (SSIM), Peak Signal-to-Noise Ratio (PSNR), and Learned Perceptual Image Patch Similarity (LPIPS) are reported to evaluate the rendering (NVS) quality.

\textbf{Baselines.}
We compare Denoising-GS with the ten most recent GS-based NVS methods, where vanilla 3DGS~\cite{kerbl20233d} and 3DGS-MCMC~\cite{kheradmand20243d} serve as the baseline methods.
These compared methods cover a wide range of recent developments, including GS with uncertainty estimation~\cite{li2024variational}, faster convergence with limited resources~\cite{mallick2024taming}, artifact suppression to improve rendering~\cite{yu2024mip}, alternative primitives (kernels) for scene representations~\cite{hamdi2024ges,li20253d}, gradient-based rendering improvements~\cite{zhang2024pixel,ye2024absgs}, and a reevaluation of the densification and pruning stages~\cite{yang2025improving}.
All baseline methods follow the original training protocol and are discussed further in the Appendix.

\subsection{Results}
\ref{tb:exp_1} shows quantitative evaluations of novel view synthesis quality. 
Across all three datasets, Denoising-GS outperforms all compared methods.

\begin{table*}[t]
\caption{The comparison of Gaussian primitives required for scene representation among the top five NVS methods and the baselines on Mip-NeRF 360 dataset, with $G$ denoting the number of required Gaussian primitives in millions (M).
The \textcolor{red!60}{red}, \textcolor{orange!60}{orange} and \textcolor{yellow}{yellow} colors represent the top three results. 
}
\centering
\resizebox{\linewidth}{!}{
\begin{tabular}{r|cccccccccccccc|cc}
\toprule [1pt]
\multirow{2}{*}{Method}&\multicolumn{2}{c}{Bicycle}&\multicolumn{2}{c}{Bonsai}&\multicolumn{2}{c}{Counter}&\multicolumn{2}{c}{Garden}&\multicolumn{2}{c}{Kitchen}&\multicolumn{2}{c}{Room}&\multicolumn{2}{c|}{Stump}&\multicolumn{2}{c}{AVG.}\\
&PSNR$\uparrow$&$G\downarrow$&PSNR$\uparrow$&$G\downarrow$&PSNR$\uparrow$&$G\downarrow$&PSNR$\uparrow$&$G\downarrow$&PSNR$\uparrow$&$G\downarrow$&PSNR$\uparrow$&$G\downarrow$&PSNR$\uparrow$&$G\downarrow$&PSNR$\uparrow$&$G\downarrow$\\
\midrule
Vanilla 3DGS&25.64&5.83&32.29&\cellcolor{yellow!40}1.25&29.08&\cellcolor{yellow!40}1.16&27.73&5.05&31.35&\cellcolor{orange!40}1.73&31.75&\cellcolor{yellow!40}1.46&26.97&\cellcolor{yellow!40}4.55&29.258&\cellcolor{yellow!40}3.004\\

3DGS-MCMC&\cellcolor{orange!40}26.15&5.90&\cellcolor{yellow!40}32.88&1.30&\cellcolor{yellow!40}29.51&1.20&\cellcolor{orange!40}28.16&5.20&\cellcolor{orange!40}32.27&\cellcolor{yellow!40}1.80&\cellcolor{orange!40}32.48&1.50&\cellcolor{red!40}27.80&4.75&\cellcolor{orange!40}29.893&3.093\\

Mip-Splatting&\cellcolor{yellow!40}25.95&7.93&32.55&1.60&29.36&1.47&\cellcolor{yellow!40}27.98&5.59&31.26&2.11&31.92&2.08&27.18&5.70&29.456&3.783\\

Pixel-GS&25.74&8.55&32.66&2.06&29.31&2.49&27.81&7.53&\cellcolor{yellow!40}32.00&3.04&31.47&2.47&27.18&6.65&29.454&4.684\\

GS-LPM&25.90&\cellcolor{yellow!40}5.75&32.35&1.31&29.21&1.21&27.77&\cellcolor{yellow!40}5.01&31.72&\cellcolor{yellow!40}1.80&31.49&1.56&\cellcolor{yellow!40}27.21&5.03&29.378&3.096\\

3D-HGS&25.69&\cellcolor{orange!40}5.51&\cellcolor{red!40}33.50&\cellcolor{red!40}1.14&\cellcolor{red!40}29.81&\cellcolor{orange!40}1.10&27.95&\cellcolor{red!40}4.49&\cellcolor{red!40}32.34&\cellcolor{red!40}1.62&\cellcolor{yellow!40}32.42&\cellcolor{orange!40}1.39&26.71&\cellcolor{orange!40}4.31&\cellcolor{yellow!40}29.772&\cellcolor{orange!40}2.794\\
\midrule
\textbf{Denoising-GS}&\cellcolor{red!40}26.17&\cellcolor{red!40}5.32&\cellcolor{orange!40}32.94&\cellcolor{orange!40}1.17&\cellcolor{orange!40}29.57&\cellcolor{red!40}1.08&\cellcolor{red!40}28.27&\cellcolor{orange!40}4.69&\cellcolor{red!40}32.34&\cellcolor{red!40}1.62&\cellcolor{red!40}32.58&\cellcolor{red!40}1.35&\cellcolor{orange!40}27.60&\cellcolor{red!40}4.29&\cellcolor{red!40}29.925&\cellcolor{red!40}2.789\\

\toprule [1pt]
\end{tabular}
}
\label{tb:exp_2}
\end{table*}

\begin{figure*}[t]
    \centering
    \includegraphics[width=\linewidth]{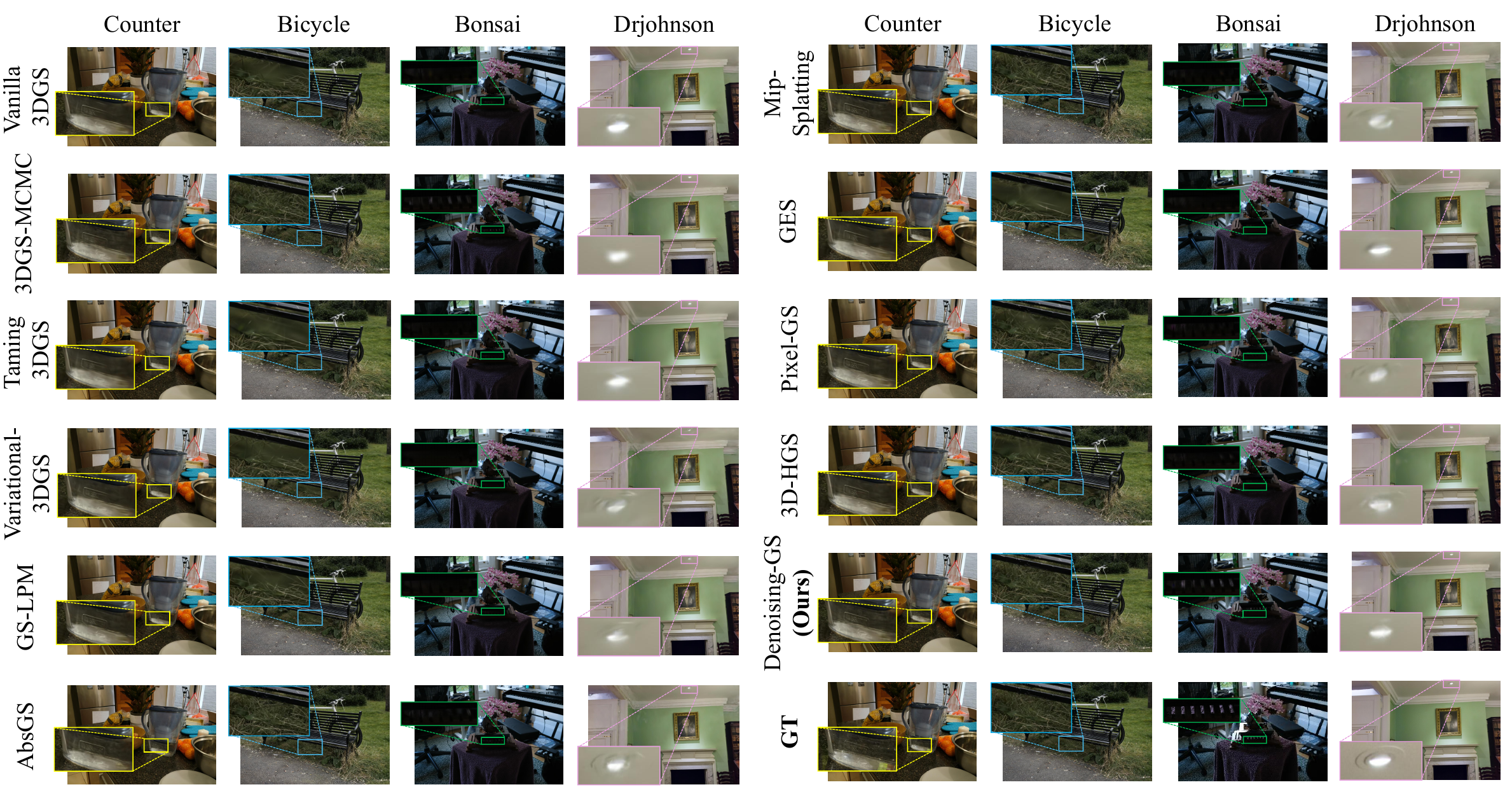}
    \caption{Visual Comparisons on Novel View Synthesis of four scenes in Mip-NeRF 360 and Deep Blending datasets. More results are provided in the Appendix.}
    \label{fig:exp}
\end{figure*}

\begin{table*}[t]
\caption{Ablation results of NVS on Mip-NeRF 360, Tank and Temples, and Deep Blending datasets. 
The top results are indicated by bold (best), underline (second best), and \textcolor{red}{red} represent the coefficients and implementation reported in \ref{sec:Experiments}.}
\vspace{-0.2cm}
\centering
\resizebox{\linewidth}{!}{
\begin{tabular}{c|ccc|cc|ccc|ccc|ccc}

\toprule [1pt]
\multirow{2}{*}{\#}&\multicolumn{3}{c|}{Coefficients}&\multirow{2}{*}{\makecell{Prune \\ Percent}}&\multirow{2}{*}{\makecell{Densify\\Percent}}&\multicolumn{3}{c|}{Mip-NeRF 360}&\multicolumn{3}{c|}{Tanks and Temples}&\multicolumn{3}{c}{Deep Blending}\\

&$\alpha$&$\beta_1$&$\beta_2$&&&SSIM$\uparrow$&PSNR$\uparrow$&LPIPS$\downarrow$&SSIM$\uparrow$&PSNR$\uparrow$&LPIPS$\downarrow$&SSIM$\uparrow$&PSNR$\uparrow$&LPIPS$\downarrow$\\ \toprule [1pt]
\multirow{7}{*}{(a)}&\cellcolor{red!40}0.05&\cellcolor{red!40}0.9&\cellcolor{red!40}0.5&\cellcolor{red!40}10\%&\cellcolor{red!40}0.05\%  &\cellcolor{red!40}\textbf{0.897}&\cellcolor{red!40}29.925&\cellcolor{red!40}\textbf{0.142}&\cellcolor{red!40}\textbf{0.868}&\cellcolor{red!40}24.561&\cellcolor{red!40}\underline{0.149}&\cellcolor{red!40}\textbf{0.906}&\cellcolor{red!40}\textbf{29.872}&\cellcolor{red!40}\underline{0.239}\\

&0.05&0.9&0.1&10\%&0.05\%  &\textbf{0.897}&\underline{29.930}&\textbf{0.142}&\underline{0.867}&\textbf{24.606}&0.150&\textbf{0.906}&\underline{29.849}&\underline{0.239}\\

&0.05&0.9&1&10\%&0.05\%  &\textbf{0.897}&29.924&\textbf{0.142}&\underline{0.867}&24.499&\underline{0.149}&\textbf{0.906}&29.836&\textbf{0.238}\\

&0.1&0.9&0.5&10\%&0.05\%  &\textbf{0.897}&29.901&\underline{0.143}&\underline{0.867}&24.458&\textbf{0.148}&\underline{0.904}&29.400&0.240\\

&0.5&0.9&0.5&10\%&0.05\%  &\underline{0.896}&29.920&\textbf{0.142}&\textbf{0.868}&\underline{24.572}&\textbf{0.148}&0.901&29.527&0.244\\

&0.05&0.5&0.5&10\%&0.05\%  &\textbf{0.897}&29.891&\textbf{0.142}&\underline{0.867}&24.486&\underline{0.149}&\textbf{0.906}&29.827&\underline{0.239}\\

&0.05&0.1&0.5&10\%&0.05\%  &\textbf{0.897}&\textbf{29.942}&\textbf{0.142}&0.866&24.456&0.150&0.903&29.493&0.240\\

\midrule
\midrule

\multirow{4}{*}{(b)}&0.05&0.9&0.5&20\%&0.05\%&\underline{0.896}&29.864&0.144&\underline{0.865}&24.496&0.154&0.901&29.120&0.243\\

&\cellcolor{red!40}0.05&\cellcolor{red!40}0.9&\cellcolor{red!40}0.5&\cellcolor{red!40}10\%&\cellcolor{red!40}0.05\%  &\cellcolor{red!40}\textbf{0.897}&\cellcolor{red!40}\textbf{29.925}&\cellcolor{red!40}\textbf{0.142}&\cellcolor{red!40}\textbf{0.868}&\cellcolor{red!40}\textbf{24.561}&\cellcolor{red!40}\textbf{0.149}&\cellcolor{red!40}\textbf{0.906}&\cellcolor{red!40}\textbf{29.872}&\cellcolor{red!40}\textbf{0.239}\\

&0.05&0.9&0.5&20\%&0.1\%&\underline{0.896}&29.866&0.144&\underline{0.865}&24.472&\underline{0.153}&\underline{0.904}&29.558&\underline{0.240}\\
&0.05&0.9&0.5&10\%&0.1\%&\textbf{0.897}&\underline{29.900}&\underline{0.143}&\textbf{0.868}&\underline{24.521}&\textbf{0.149}&0.905&\underline{29.700}&\textbf{0.239}\\

\toprule [1pt]
\end{tabular}
}

\label{tb:ab1}
\end{table*}

\textbf{Primitive Efficiency.}
In scene representation, there is a trade-off between rendering quality and the number of primitives used.
As shown in \ref{tb:exp_1}, Denoising-GS outperforms methods~\cite{li2024variational,mallick2024taming,hamdi2024ges,ye2024absgs,kerbl20233d} in terms of NVS quality on the Mip-NeRF 360 dataset, while methods~\cite{kheradmand20243d,yu2024mip,yang2025improving,zhang2024pixel,li20253d} achieve similar quantitative results.
To further examine the efficiency of scene representation, \ref{tb:exp_2} compares the number of primitives used by the top five comparison methods~\cite{kheradmand20243d,yu2024mip,yang2025improving,zhang2024pixel,li20253d} (based on PSNR results) and the vanilla 3DGS~\cite{kerbl20233d} on the Mip-NeRF 360 dataset.
Specifically, Denoising-GS achieves the highest NVS performance while requiring the fewest primitives.
3D-HGS~\cite{li20253d} employs nearly the same number of primitives as Denoising-GS but yields suboptimal NVS results. 
Additionally, Denoising-GS saves 10\% or more primitives compared with~\cite{kerbl20233d,kheradmand20243d,yang2025improving}, whereas \cite{yu2024mip,zhang2024pixel} use significantly more primitives than Denoising-GS.

\textbf{Visualizing Comparisons.} 
\ref{fig:exp} shows comparisons across four scenes (\textit{Counter}, \textit{Bicycle}, \textit{Bonsai}, and \textit{Drjohnson}) from~\cite{barron2022mip,hedman2018deep}.
More scenes will be discussed in the Appendix.
Denoising-GS provides better visual quality than all ten other methods.
Specifically, \textit{Counter} is a well-lit indoor scene featuring semi-transparent objects and metallic specular surfaces. 
\textit{Bicycle} is a large-scale outdoor scene featuring extensive areas of dense and finely detailed grass.
\textit{Bonsai} is a dimly-lit indoor environment where motion blur occurs due to camera shake during capture. 
\textit{Drjohnson} is a complex indoor scene with multiple light sources, presenting substantial challenges for NVS under such intricate illumination conditions. 
The analysis below provides a detailed comparison of the four scenes.
\textit{Counter}: We evaluate the NVS capability for the semi-transparent brand text on the water bottle (yellow box). While \cite{kheradmand20243d,mallick2024taming,hamdi2024ges} fails to render this region clearly, Denoising-GS recovers sharper details.
\textit{Bicycle}: The NVS of weeds beneath the seat (blue box) is examined. Because this area is invisible in most of the viewpoints, \cite{kerbl20233d,mallick2024taming,li2024variational,yang2025improving,hamdi2024ges,zhang2024pixel,li20253d} produce noticeable floating artifacts. 
In contrast, with the Uncertainty-based Denoising and the Spatial Coherence Refinement, Denoising-GS produces clear rendering results while using fewer primitives (5.32M) than 3DGS-MCMC~\cite{kerbl20233d} (5.90M), AbsGS~\cite{ye2024absgs} (6.44M), and Mip-splatting~\cite{yu2024mip} (7.93M). 
\textit{Bonsai}: Rendering the structure beneath the flowerpot (green box) is particularly challenging because of severe illumination insufficiency. 
While almost all comparing methods hardly recover any structural details, Denoising-GS successfully reconstructs plausible geometry and texture in novel views.
\textit{Drjohnson}: We analyze the upper lighting area (pink box), where specular effects pose persistent challenges. 
Most comparison methods exhibit floating artifacts in this area, while Denoising-GS produces lighting reconstruction without artifacts. 
Moreover, our method maintains high rendering quality in other areas of the room outside the highlighted region.

\section{Ablation Studies}\label{sec:ablations}

Since the three benchmarks differ in characteristics such as image resolution and the quality of input SfM point clouds, our ablation aims to confirm the effectiveness of the proposed model across all scenes. 
Accordingly, we conduct ablations across all three datasets and select the hyperparameters that perform best on average, rather than optimizing separate settings for each dataset.

\begin{figure*}[t]
    \centering
    \includegraphics[width=\linewidth]{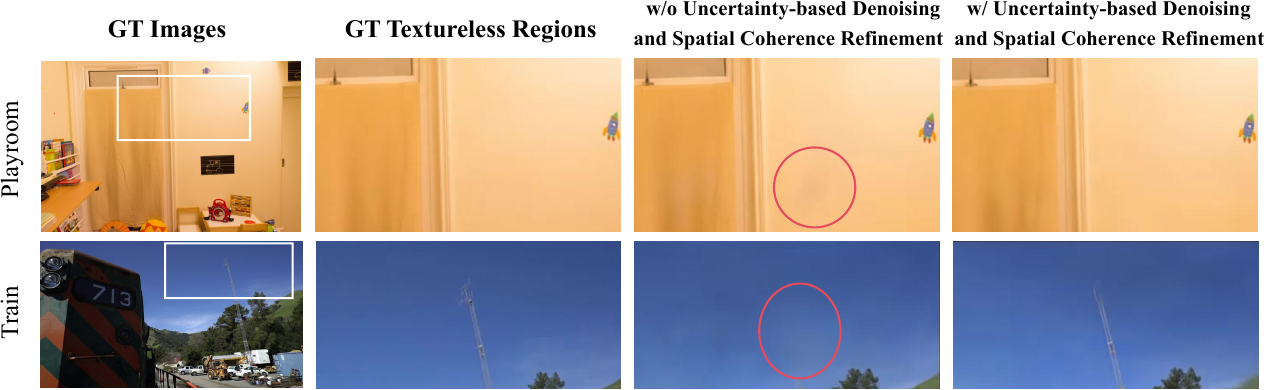}
    \caption{Visual ablation of NVS in textureless regions, demonstrating the contributions of Uncertainty-based Denoising and Spatial Coherence Refinement.}
    \label{fig:ab1}
\end{figure*}

\textbf{Controlling Coefficients.} 

\ref{tb:ab1}(a) presents ablation results for three 
coefficients in \ref{sec:4.2} and \ref{sec:4.3}. 
The parameter $\alpha$ controls the strength of correction applied to the momentum term relative to the exploration noise. A larger correction often slows down exploration, which in turn affects convergence. 
Therefore, we set $\alpha$ to a relatively small value, and experiments show that $\alpha = 0.05$ achieves the best and most balanced performance among the three datasets. 
The coefficient $\beta_1$ controls how smooth the momentum update is. 
Since the convergence behaviors differ between indoor and outdoor scenes, the different weights assigned to historical momentum and current gradients do not perform consistently across various datasets. 
We therefore select $\beta_1 = 0.9$, which achieves stable performance across all three datasets and matches the commonly used smoothing factor in momentum-biased optimization.
The coefficient $\beta_2$ controls the strength of the denoising term. 
We set $\beta_2 = 0.5$ to achieve the best balance between denoising effectiveness and convergence speed in both indoor and outdoor scenes.

\textbf{Uncertainty-based Denoising and Spatial Coherence Refinement.}
It is demonstrated~\cite{hanson2025pup} that removing too many primitives, while sometimes able to maintain visual quality, results in a significant decline in quantitative metrics. 
Conversely, pruning only a few primitives keeps rendering quality high but lowers the effectiveness of the denoising and pruning stages. To investigate this trade-off, we remove 10\% and 20\% of primitives, which corresponds to the removal of approximately 310K and 620K primitives, respectively, from the Mip-NeRF 360 dataset. 
At the same time, our Spatial Coherence Refinement strategy aims to improve scene representation while adding as few new points as possible. We examine densification rates of 0.05\% and 0.1\%, which correspond to roughly 5.6K and 13.9K additional points on Mip-NeRF 360 under the 10\% pruning setting from the previous denoising stage. As shown in \ref{tb:ab1}(b), pruning 10\% of primitives combined with a 0.05\% densification rate (an average reduction of 0.3M primitives per scene) achieves the best balance between compactness and rendering fidelity.
Additionally, in \ref{fig:ab1}, we present a visual ablation of the Uncertainty-based Denoising and Spatial Coherence Refinement strategies, demonstrating that our method provides a superior representation capability in textureless regions.

\section{Conclusion}\label{sec:conclusion}

In this paper, we introduce a new perspective on GS optimization as a primitive denoising process and propose Denoising-GS, a spatial-aware denoising framework tailored for Gaussian primitives.
We first introduce the momentum-biased stochastic exploration to allow primitives to remember the spatial process of optimization. 
On top of the momentum-biased optimizer, we design the Spatial Gradient-based Denoising strategy and the Uncertainty-based Denoising module for an accurate and compact scene representation.
Finally, the Spatial Coherence Refinement strategy divides primitives in sparse areas to enhance the quality of the scene representation.

\textbf{Limitations and Future Work.}
Rendering quality and the number of primitives involve a trade-off. In regions with sparse texture, such as large uniform areas of sky or low-texture walls, our method may exhibit a slight decline in rendering quality because of the denoising (pruning) of Gaussian primitives.
For future work, we plan to extend its application to flattened primitives and investigate the potential for mesh extraction.



{
    \small
    \bibliographystyle{elsarticle-num}
    \bibliography{ref}
}

\end{document}